\documentclass[10pt,twocolumn,twoside]{IEEEtran}
\usepackage{mathrsfs}
\usepackage{amssymb}
\usepackage{algorithm}
\usepackage{algorithmic}
\usepackage{{inputenc}}
\usepackage{balance}
\usepackage{amsmath}
\usepackage{multirow,tabularx}
\usepackage{{inputenc}}
\usepackage{balance}
\usepackage{bm}
\usepackage{graphicx}
\usepackage{hyperref}
\usepackage{breakurl}
\usepackage{subfig}

\begin{document}

\title{Learning Cross Space Mapping via DNN using Large Scale Clickthrough Data}

\author{Wei~Yu$^{*}$,
        Kuiyuan~Yang,
        Yalong~Bai,
        Hongxu~Yao,
        and~Yong~Rui,~\IEEEmembership{Fellow,~IEEE}

\thanks{W. Yu, Harbin Institute of Technology, Harbin, Heilongjiang 150090. E-mail:w.yu@hit.edu.cn}
\thanks{K. Yang, Microsoft Research, Beijing, Beijing 100080. E-mail: kuyang@microsoft.com.}
\thanks{Y. Bai, Harbin Institute of Technology, Harbin, Heilongjiang 150090. E-mail:ylbai@mtlab.hit.edu.cn}
\thanks{H. Yao, Harbin Institute of Technology, Harbin, Heilongjiang 150090. E-mail:h.yao@hit.edu.cn}
\thanks{Y. Rui, Microsoft Research, Beijing, Beijing 100080. E-mail: yongrui@microsoft.com.}}

\markboth{IEEE Transactions on Multimedia,~Vol.~X,
No.~XX,~Month~Year}{} \maketitle

\begin{abstract}
The semantic gap between low-level image pixels and high-level semantics has been progressively bridged by the continuously developing Deep Neural Network (DNN), and achieved great success in almost all image classification tasks.
To extend the power of DNN to image retrieval task, we proposed an unified DNN model for image-query similarity calculation by simultaneously modeling image and query in one network.
The unified DNN is named as Cross Space Mapping (CSM) model, which contains two main parts, i.e., convolutional part and query-embedding part.
The image and query are mapped to a common vector space via these two parts respectively, and image-query similarity is naturally defined as inner product of their mappings in the space. To ensure good generalization ability of the DNN, we learn the weights of the DNN on a large scale clickthrough dataset which consists of 23 million clicked image-query pairs between 1 million images and 11.7 million queries.
Both the qualitative results and quantitative results on an image retrieval evaluation task with 1000 queries demonstrate the superiority of the proposed method.
\end{abstract}


%
\IEEEpeerreviewmaketitle

\section{Introduction}
With the popularization of digital cameras and storage devices, millions images are taken everyday and billion images are hosted in photo-sharing websites and image search engines.
A nature problem with such gigantic image collections is how to retrieve the relevant images for everyday users, which is also well known as image retrieval problem.
Though image retrieval is with similar user-interaction mode with document retrieval (users provide a few keywords as query, and the machine returns a list of relevant documents), image retrieval is more challenge as machine cannot directly use string matching to check whether the textual query matching with the candidate images.
Current image search engines mainly rely on the surrounding texts of an image to represent textual information conveyed in the image, and convert image retrieval into document retrieval.
However, surrounding texts are not always available or relevant to the image, which leads large number of images irretrievable or irrelevant.

In order to make all images retrievable and improve the relevance of retrieved images, the machine needs the ability to directly measure the image-query similarity by extracting information from image itself. Though sounds intuitive, this is a difficult task and far from being solved for the following two reasons:
 \begin{itemize}
   \item Extract semantic information from images is hard even with the state-of-the-art hand crafted image features (e.g., super-vector coding~\cite{SVcode}, fisher vector~\cite{fisherKernel}, spatial pyramid matching~\cite{SPM}, etc.).
   \item The number of possible queries is huge even if not infinite, it is impractical to build classifiers query by query as image classification tasks.
 \end{itemize}

\begin{figure}[ht]
\centering
\includegraphics[width=0.45\textwidth,page=1]{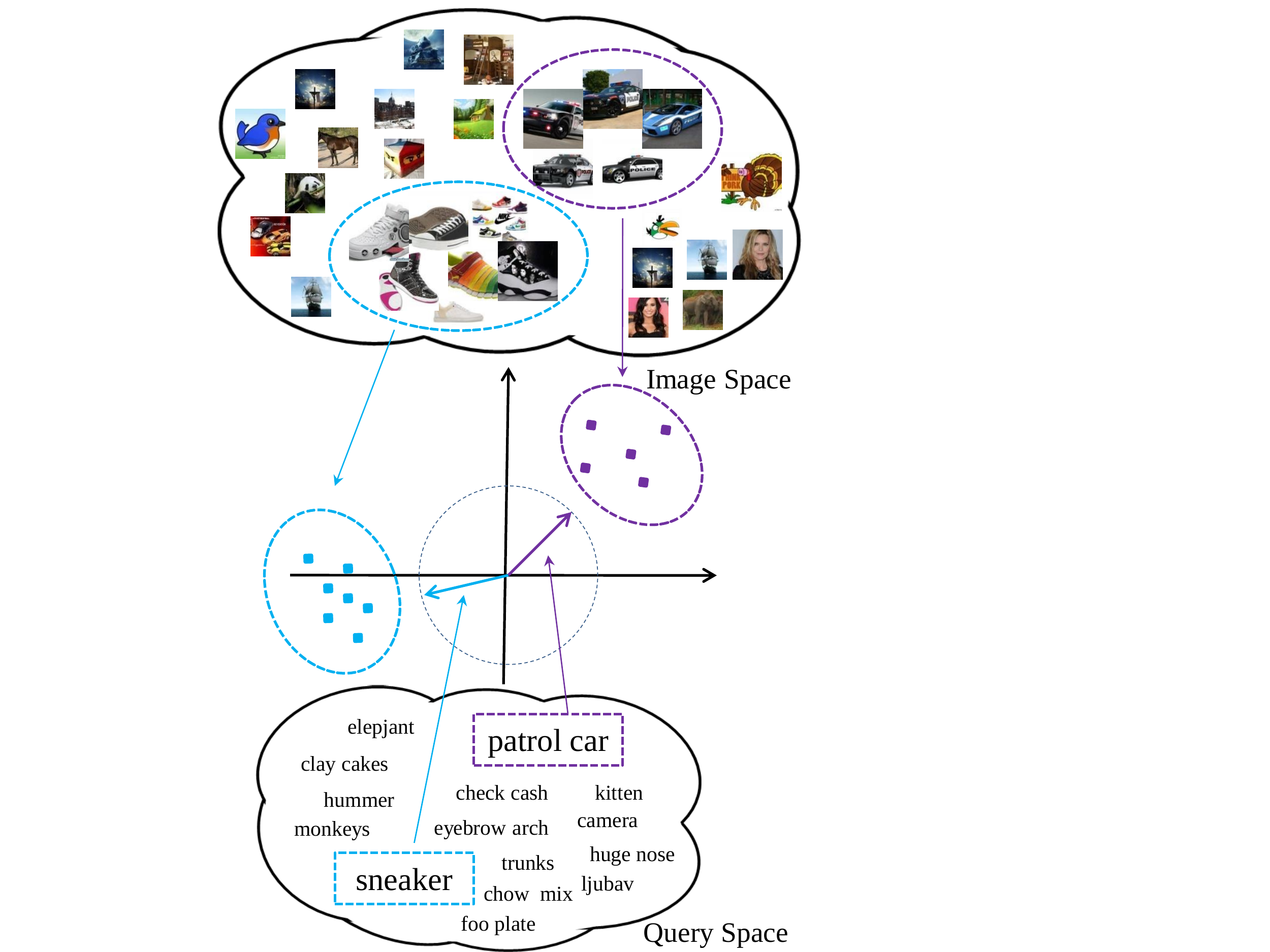}
\caption{The cross space mapping model. Both image space and query space are mapped to a common space, where images are represented as visual feature vectors and queries are represented as textual weight vectors.}
\label{CSMframe}
\end{figure}

Recent significant progress in DNN has shown the possibility and superiority in automatically learning representations from raw inputs such as images and texts. Inspired by the success of DNN in image classification and word embedding tasks, we proposed an unified DNN to model the image-query similarity. The proposed DNN unifies Convolutional Neural Network (CNN) and Word Embedding Network (WEN) to generate representations from images and queries respectively, where the final outputs of CNN and WEN are residing in the same vector space and their inner product is defined as the image-query similarity. CNN has shown its superiority over hand crafted image features in extracting semantic information from images via the automatically learned features~\cite{krizhevsky2012imagenet,ZF}. WEN has been successfully used in natural language processing tasks by learning low dimensional vector representations of words~\cite{WordEmbed}, and query representation is modeled by the linear weighted combination of word vectors.
With the unified DNN, both image and query are mapped into the same feature vector space as illustrated in Figure~\ref{CSMframe}.

DNN requires large number of training data to learn its parameters.
Here, we utilize a large scale clickthrough dataset collected from Bing image search as the training dataset, which contains of 23 million clicked image-query pairs from 11.7 million queries and 1 million images~\cite{Clickage}. The large number of queries, images and their associations provide a well coverage of the sample space. With such large number training examples, there is no observable overfitting problem even without using dropout~\cite{krizhevsky2012imagenet}.

Qualitative results show our learned CSM model constructs a meaningful common vector space for both image and query.  We further evaluate the learned DNN on an image retrieval task with 1000 queries. The quantitative results on image retrieval comparing several competing methods demonstrate the effectiveness of the proposed method.

The rest of the paper is organized as follows. Related work is presented in Section~\ref{s:relatedWork}, the unified DNN for jointly image-query modeling and learning is introduced in Section~\ref{CSM}. Experimental results on a large scale clickthrough dataset are presented in Section~\ref{experiment}. Finally, we conclude this work in Section~\ref{Conclusion}.

\section{RELATED WORK}
\label{s:relatedWork}
As one important domain of information retrieval, image retrieval has been intensively studied for decades in both academic and industrial community~\cite{ImageSearch20}. However, current image retrieval systems still mainly rely on surrounding texts of images to perform the retrieval task. As the missing and noisy problem of surrounding texts, many research works have been proposed to use the image content to measure image-query similarity.
With the continuously developing image content understanding techniques especially with the rebirth of convolutional neural network, the image content is gradually playing more important role.

\subsection{Image annotation as intermediate step}
Automatic image annotation is the process by which a machine automatically assigns keywords to an image, and image retrieval is performed over the annotated keywords. A typical pipeline of image annotation is firstly representing images with visual features, then predicting the keywords of images by machine learning algorithms. According the algorithms used, image annotation approaches can be roughly divided into two categories, i.e., model-based approach~\cite{vogel2004natural,tieu2004boosting} and data-driven approach~\cite{annoSearch}.

In model-based approaches, image annotation is performed as multi-class or multi-label classification problem, where a manually labeled dataset is used for learning models such as SVM and boosting~\cite{vogel2004natural,tieu2004boosting}. Model-based approaches often work with thousands categories, which are impractical to scale up to millions or even more queries.

Compared with model-based approaches, data-driven approaches can be performed without the limits of the number of queries.
In a data-driven approach, the annotation of an image is assigned by propagating annotations of its similar images~\cite{annoSearch}. Due the limitation of low-level image features, data-driven approaches only work well for images with enough duplicates in the training set. It is worth mentioning that image annotation is performed in image retrieval as an intermediate step, and queries are further needed to compare with the annotations to accomplish the retrieval.

\subsection{Joint image and query modeling}
To avoid the intermediate step of image annotation, many works studied how to jointly modeling image and query where image-query similarity is directly estimated. There are two main directions in this area, one is using generative models and another is using discriminative models.

Generative models are applied widely in joint image and query modeling as they are easy to take different modalities into account. Different kinds of generative models have been proposed for joint image and query modeling, including latent Dirichlet allocation~\cite{barnard2003matching}, probabilistic latent semantic analysis~\cite{monay2004plsa}, hierarchical Dirichlet processes~\cite{yakhnenko2008annotating}, machine translation methods~\cite{duygulu2006object} and deep Boltzmann machine~\cite{Multimodal_DBM}, etc.
As it is still difficult to learn probability on raw images, hand crafted image features are used in the modeling.

Discriminative models are generally with better performance. In discriminative models, joint kernels over image and query are defined and learned for ranking images~\cite{grangier2008discriminative,wsabie}. Though different image features and diverse kernel functions are considered in these works, their modeling ability still limited by the visual features and their shallow structures. In~\cite{DeViSE}, deep visual-semantic embedding model is proposed to measure image-query similarity by automatically learned convolutional neural network. Unlike our methods, the method still requires ImageNet to do supervised pretraining.

\section{CROSS SPACE MAPPING MODEL}
\label{CSM}

\begin{figure*}[t]
\centering
\includegraphics[width=0.98\textwidth,page=1]{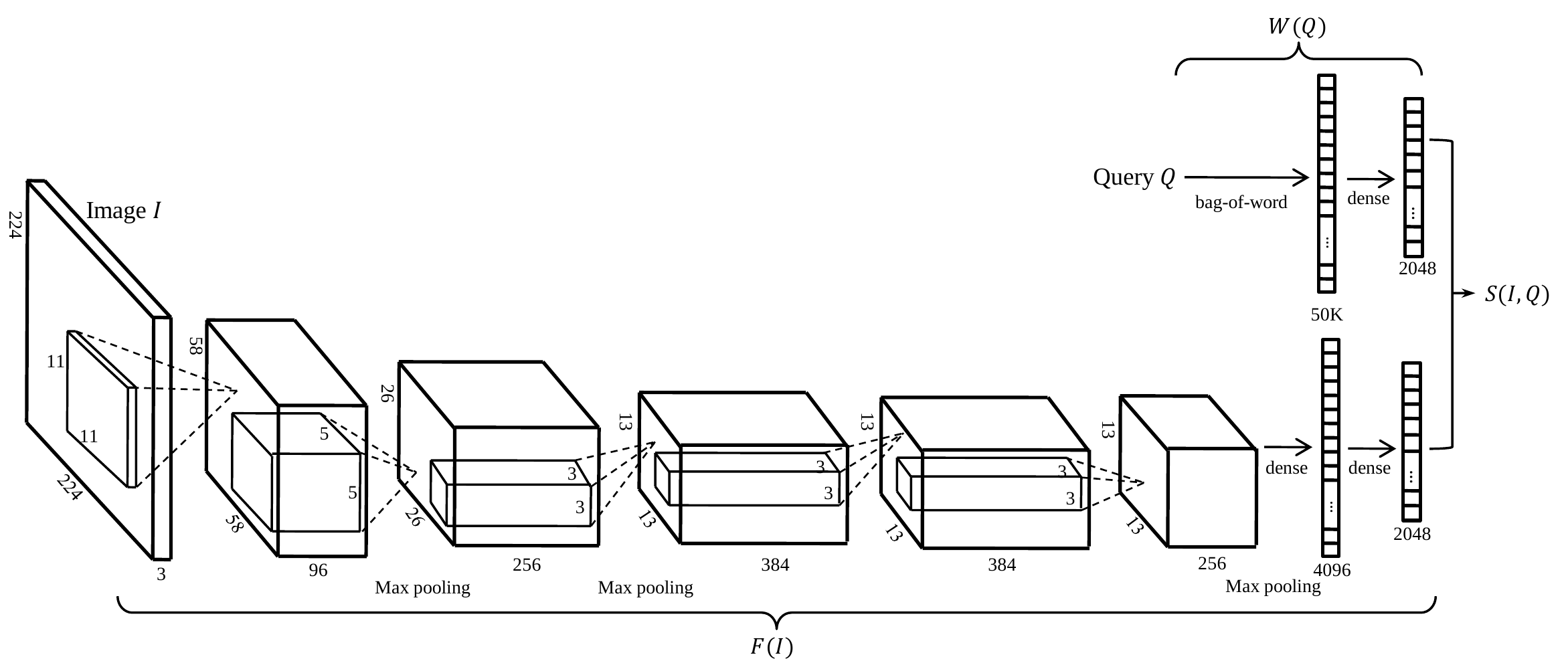}
\caption{Architecture of the unified DNN. The upper part is WEN modeling textual query mapping  while the lower part is CNN modeling image mapping.}
\label{DNNframe}
\end{figure*}

In this section, the unified DNN is described for image-query similarity modeling and accomplish the cross space mapping. We first introduce CNN and WEN separately for image and query modeling, and then unify these two networks together into one unified DNN to define the image-query similarity. Then the training procedure is introduced to learn the DNN model parameters.

\subsection{CNN for Image Modeling}
Images are stored as raw pixels in the machine, we use standard CNN~\cite{krizhevsky2012imagenet,lecun1989backpropagation} without softmax outputs for image modeling. The CNN contains seven layers with weights, including five convolutional layers and two fully-connected layers, three max-pooling layers are used following the first, second and fifth convolutional layers, two local contrast normalization layers are used following the first and second max-pooling layers. More details of these operations can be referred in~\cite{krizhevsky2012imagenet}. The lower part of Figure~\ref{DNNframe} illustrates the architecture of the image part. Via the CNN, an image $I$ is mapped to a $d$-dimensional vector space and denoted as $F(I)\in\mathbb{R}^d$.

\subsection{WEN for Query Modeling}
Queries are stored as a set of words in the machine, word embedding~\cite{WordEmbed} is leveraged for query modeling.
To this end, we build a vocabulary $\mathcal{V}$ formed by 50K words with top word-frequency in training set, where $|\mathcal{V}|$=50K.
With word embedding, a word $t\in\mathcal{V}$ is mapped into $d$-dimensional space as $\bold{w}(t)\in\mathbb{R}^d$ using a lookup table, and will be learned in the training procedure. Then a query $Q$ is mapped to the same space as $W(Q)\in\mathbb{R}^d$ by weighted linear combination of its words' vectors, i.e.,
\begin{equation}
W(Q)=\frac{1}{|Q|}\sum_{t\in Q} \omega (t)\bold{w}(t)
\end{equation}
where $\omega(t)$ is weights for word $t$, and normally defined as the \emph{normalized idf weighting}
\begin{equation}
\omega(t)=\frac{idf_t}{\sqrt{\sum_{t\in Q} idf_t^2}}
\end{equation}
where $idf_t=-\log(r_t)$, $r_t$ refers to the fraction of corpus queries containing the word $t$.
The upper part of Figure~\ref{DNNframe} illustrates the architecture of the query part, which is a networks with two layers, the first layer takes bag-of-words representation of query as input, the second layer outputs the query embedding vector. The parameters for word embedding is represented as the weights between the two fully-connected layers.

\subsection{Image-Query Similarity}
With the image mapping $F(I)$ and query mapping $W(Q)$, images and queries both are mapped into a common feature space, and the image-query similarity can be defined as their inner product, i.e.,
\begin{equation}
S(I,Q)=<F(I),W(Q)>~\label{IQ}
\end{equation}
where $S(I,Q)$ is image-query similarity.

As the output of the unified DNN model, $S(I,Q)$ can be used to determine whether image $I$ and $Q$ is relevant or not, and can be naturally used to ranking candidate images for a specific query.

\subsection{Training Data Preparing}
Given a clickthrough dataset denoted by $\{\mathcal{I},\mathcal{Q},M\}$, $\mathcal{I}$ and $\mathcal{Q}$ refer to the image set and query set respectively, $M$ is the click matrix which represents the corresponding clicks between images and queries in training set.
With image-query similarity $S(I,Q)$, we further define the following constraint that requires clicked image-query pairs are with large similarity:
\begin{equation}
S(I^+,Q)>S(I^-,Q)\; \; \; \; for\;  \forall I^+\in \mathcal{I}_Q\; and\; \forall I^-\in \mathcal{N}_Q
\end{equation}
where $Q\in \mathcal{Q}$, $\mathcal{I}_Q=\{I:M(I,Q)=1,I\in\mathcal{I}\}$  is the clicked images of query $Q$, $\mathcal{N}_Q=\mathcal{I}-\mathcal{I}_Q$ is unclicked images of query $Q$.

\begin{figure}[t]
\centering
\includegraphics[width=0.48\textwidth,page=1]{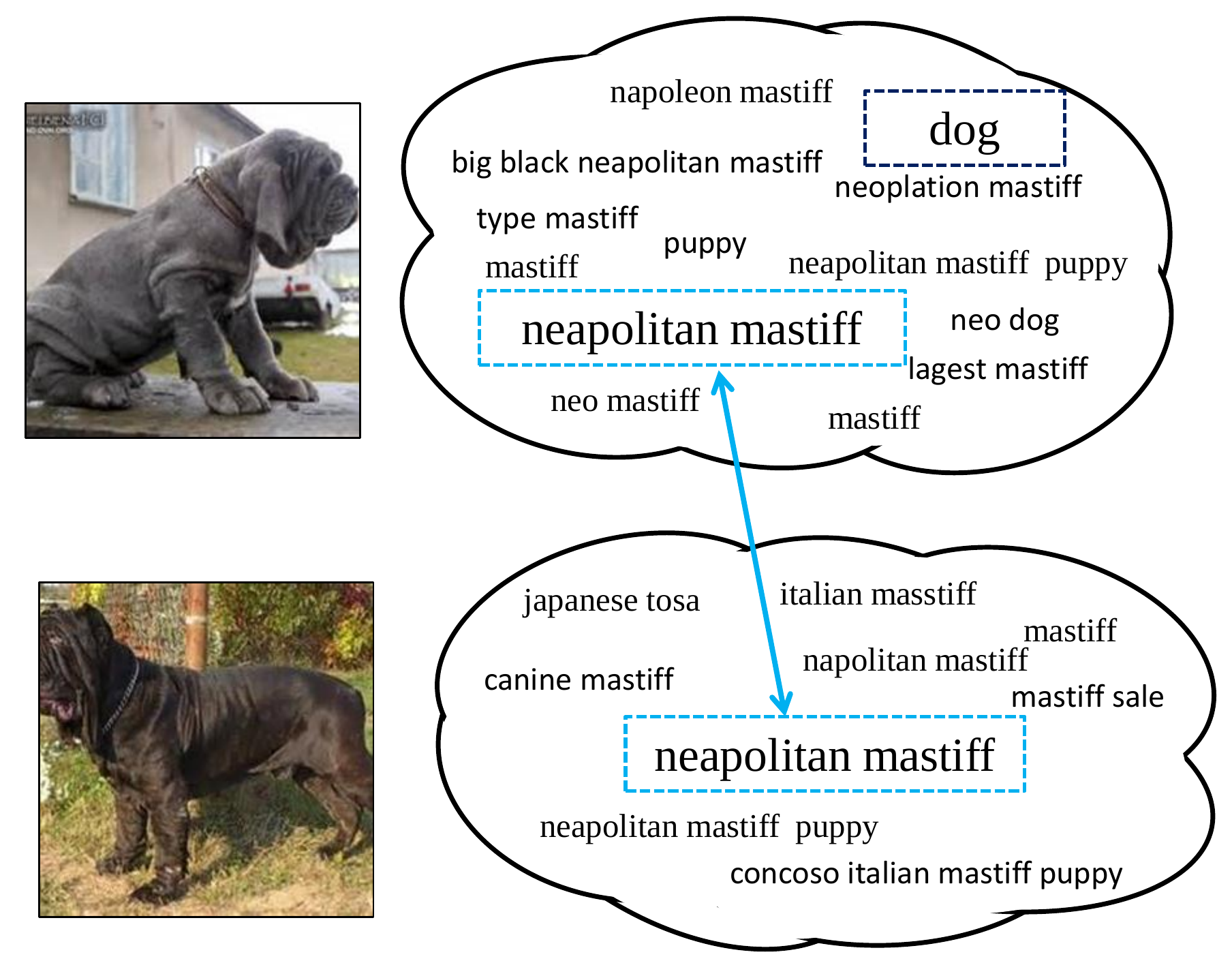}
\caption{The images sharing same queries.}
\label{imageCorresponding}
\end{figure}

In web-scale image set, the unclicked image set for each query is often too large for direct optimization.
Thus, the practical negative set $\mathcal{N}_Q$ is a subset sampled from the complementary set of $\mathcal{I}_Q$.
Here, We propose a preprocessing stage, which attempt to sample the better negative examples as negative set.
Considering the click matrix $M$ is only partially observed, that is the non-clicked image-query pairs are not necessary irrelevant.
As illustrated in Figure~\ref{imageCorresponding}, the bottom image is denoted as irrelevant to \emph{dog} by M which is actually relevant.
Yet, the top and bottom images share other same queries, such as \emph{neapolitan mastiff}, which means the bottom image should be removed from the negative set of query \emph{dog}.

Based on this idea, we denote the first order image relationship matrix as $M_{1} = M\cdot M^{T}$, and the $n^{th}$ order image relationship matrix can be defined as $M_{n}=(M_{1})^{n}$.
In this paper, we utilize $M_{2}$ to remove the potential relevant images from the negative set of specific query, and the final $\mathcal{N}_{Q}$ is sampled from the set $\{I: M(I,Q)=0 \;\&\; M_{2}(I^{+},I)=0, I^{+}\in \mathcal{I}_{Q}\}$.

\subsection{Training Objective}
In order to measure discrimination between $\mathcal{I}_Q$ and $\mathcal{N}_Q$, we define the inter-class scatter of query $Q$ as:
\begin{equation}
\begin{split}
for \;\; \forall I^+\in \mathcal{I}_Q,&\;\;\forall I^-\in \mathcal{N}_Q\\
m(Q;\theta_i,\theta_q) &= \text{min} \;  S(I^+,Q)-S(I^-,Q)\\
&=\text{min} \; W(Q)\cdot F(I^+)-W(Q)\cdot F(I^-) \\
&=\text{min} \; W(Q)\cdot (F(I^+)-F(I^-)) \\
\end{split}
\end{equation}
where $\theta_i$ and $\theta_q$ are the parameters of image mapping and query mapping respectively. As the minimum score difference of all positive-negative image pairs for query $Q$, $m(Q;\theta_i,\theta_q)$ can be regarded as margin in classification tasks, where larger margin would yield better discrimination.

\begin{figure}[t]
\centering
\includegraphics[width=0.48\textwidth,page=1]{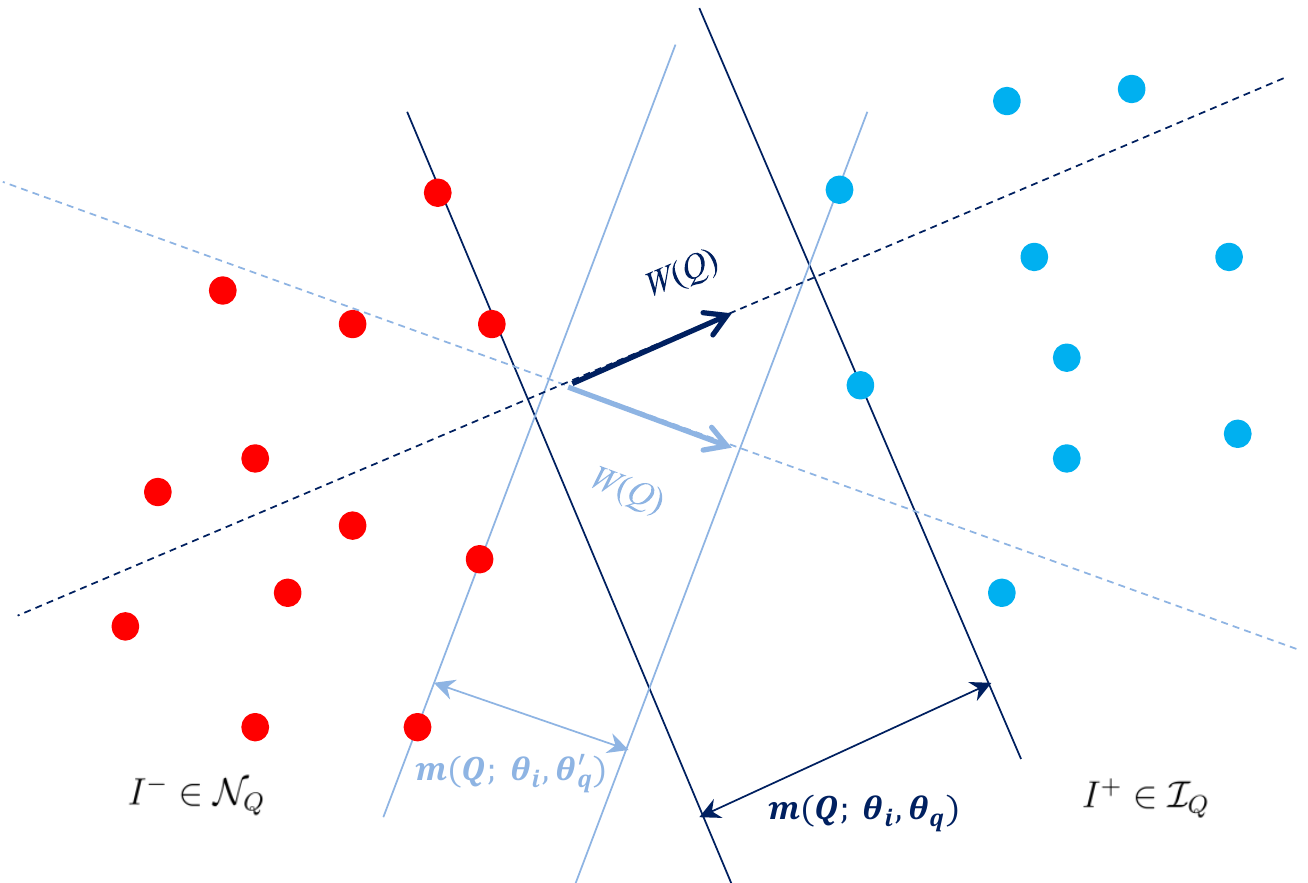}
\caption{Margins generated based on different textual query mapping.}
\label{margin}
\end{figure}

Obviously, parameter vectors $\theta_i$ and $\theta_q$ jointly determine the margin given the data. Figure~\ref{margin} shows different margins using different query mappings with fixed image mapping.
Actually, Figure~\ref{margin} could also be regarded as the cases where the visual features are preselected in image retrieval. The goal of these cases is to find the optimal query weight vector which maximizes the margin, while the visual feature has been fixed such as SIFT and GIST.

However, the preselected visual feature mapping may not have enough ability to distinguish negative set and positive set. Therefore, the inter-class scatter is also influenced by image mapping. Here, our CSM aims at learning the parameter vectors $\theta_i$ and $\theta_q$ simultaneously by enlarging the margins over all queries. The training objective of CSM could be formulated as follows:
\begin{equation}
[\theta_i^*,\theta_q^*]=\arg\; \text{max}\sum_{Q\in \mathcal{Q}} m(Q;\theta_i,\theta_q)
\end{equation}
To avoid trivial solution, both the norm of $\theta_i$ and $\theta_q$ are constrained to be less than 1.

The DNN is trained by stochastic gradient descent with a batch size of 128 queries, for each query, the loss is defined as the negative of its margin.
In the training process, the update rule for network parameters $\theta  = \{\theta_i , \theta_q\}$ is formulated as:
\begin{equation}
\begin{cases}
 &v_{t+1}=\alpha \cdot v_t - \beta \cdot \varepsilon \cdot \theta_t - \varepsilon \cdot \left \langle \frac{\partial L}{\partial \theta}|_{\theta_t} \right \rangle_{D_{t}} \\
 &\theta_{t+1}=\theta_t+v_{t+1}
\end{cases}
\end{equation}
where $t$ is the iteration index, $\alpha$ is the momentum set to 0.9, $v$ is the momentum variable, $\beta$ is the weight decay set to $10^{-5}$, $\varepsilon$ is the learning rate and $\left \langle \frac{\partial L}{\partial \theta}|_{\theta_i} \right \rangle_{D_{t}}$ is the average over the $t^{th}$ batch $D_t$ of the derivative of the objective with respect to $\theta$, evaluated at $\theta_t$. The learning rate is initially set to 0.01, and decreased by a factor of 10 when the margin on a validation set stopped improving.

\subsection{Image Retrieval System}

\begin{figure}[ht]
\centering
\includegraphics[width=0.48\textwidth,page=1]{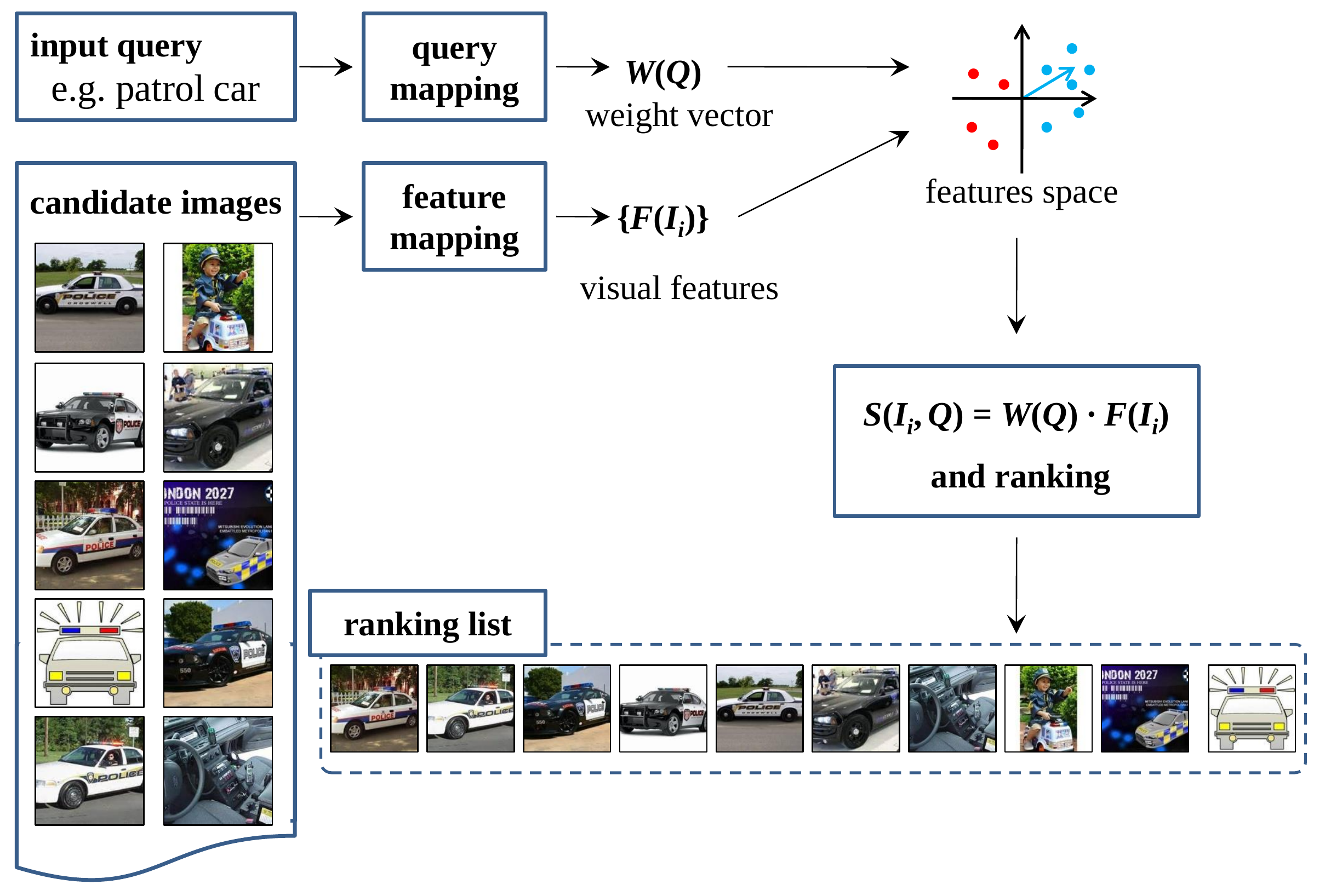}
\caption{Framework of CSM based image retrieval system.}\label{system}
\end{figure}

Based on the learned CSM, we can build a textual query based image retrieval system, as shown in Figure~\ref{system}. The candidate images in database are translated as visual feature vectors in the mapped space, and the textual query input by user is translated as weight vector in the common mapped space. Through calculating and sorting the scores of input query and candidate image pairs, this system output a ranked image list as the retrieval result.
In particular, few input queries can not be matched within the training word set, since the input words don't appear in the selected training words with top word-frequency.
In this case, retrieval system will return a random ranking result.

\section{EXPERIMENT SETTING}
\subsection{Dataset Description}
The user-click data was collected from Bing search engine, which is publicly accessible as MSR-Bing Image Retrieval Challenge~\cite{Clickage}. In this dataset, images are collected from the Web and labels are the input textual queries from Bing's users. The dataset is collected based on queries received at Bing Image Search in EN-US market. The dataset comprises two parts: the training dataset, and the dev dataset which label is judged by annotators and used as test dataset.

\begin{figure}[ht]
\centering
\includegraphics[width=0.45\textwidth,page=1]{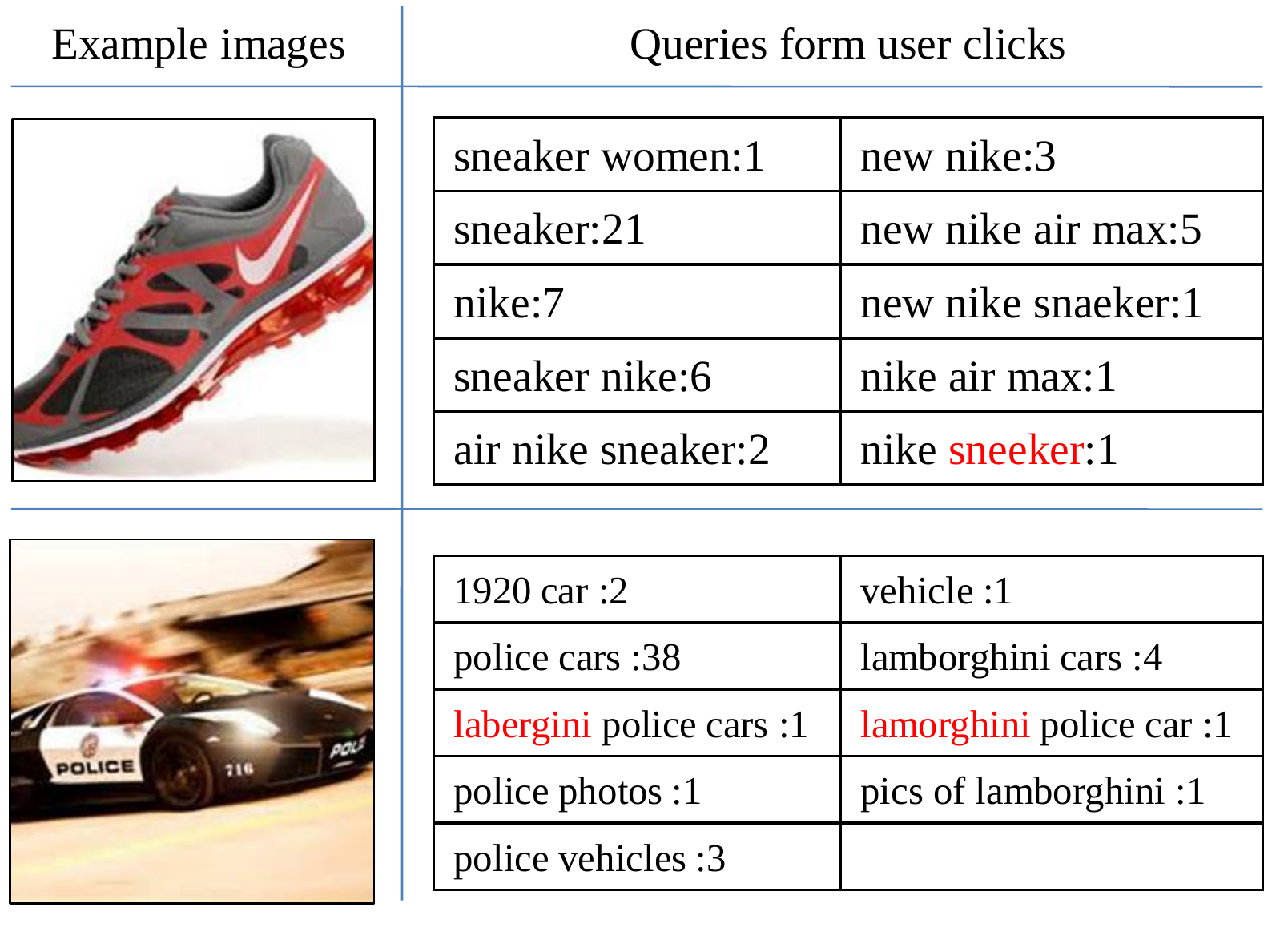}
\caption{Examples of training dataset, words marked with red font are typos by users.}\label{Train}
\end{figure}

The training dataset includes 11,701,890 queries, 1,000,000 images and 23,094,592 clicked $<$query, image$>$ pairs, where the whole clicked data is randomly sampled from one year's of Bing Image Search log. The topics of queries are wide and diverse, some examples are shown in Figure~\ref{Train}.

The test dataset is comprised of {1,000} queries and {79,665} images, which are also randomly sampled  from the one year's Bing Image Search log in EN-US market. In order to measure the relevance, a large set of plausible retrieval results are judged manually for each query.  The relevance of images are measured with three levels with respect to a query, that is \textit{Excellent = 3, Good = 2, Bad = 0}. The judgment guidelines and procedure are established to ensure high data quality and consistency.

\subsection{Evaluation Criterion}
\label{evaCr}
In order to measure the performance of the search results, we adopt Discounted Cumulated Gain (DCG) measurement to quantify the retrieval performance. The standard DCG is defined as:
\begin{equation}
DCG_n=\gamma\sum_{i=1}^{n}\frac{2^{rel_{i}}-1}{log_2(i+1)}
\end{equation}
where $n$ is the count of images in searching list, $rel_i$ is the relevance level of the result at position $i$. In our experiment, $rel_i  = {0, 2, 3}$ as previous mentioned and $n = 25$, $\gamma=0.1757$ is the normalizer that make the best $\text{DCG}_{25}$ equals to 1.

\section{EXPERIMENT RESULTS}
\label{experiment}
In this section, we demonstrate CSM based image retrieval from both qualitative results and quantitative results.

\subsection{The Learned Mapping Space}
\begin{figure*}[t]
\centering
\includegraphics[width=0.8\textwidth,page=1]{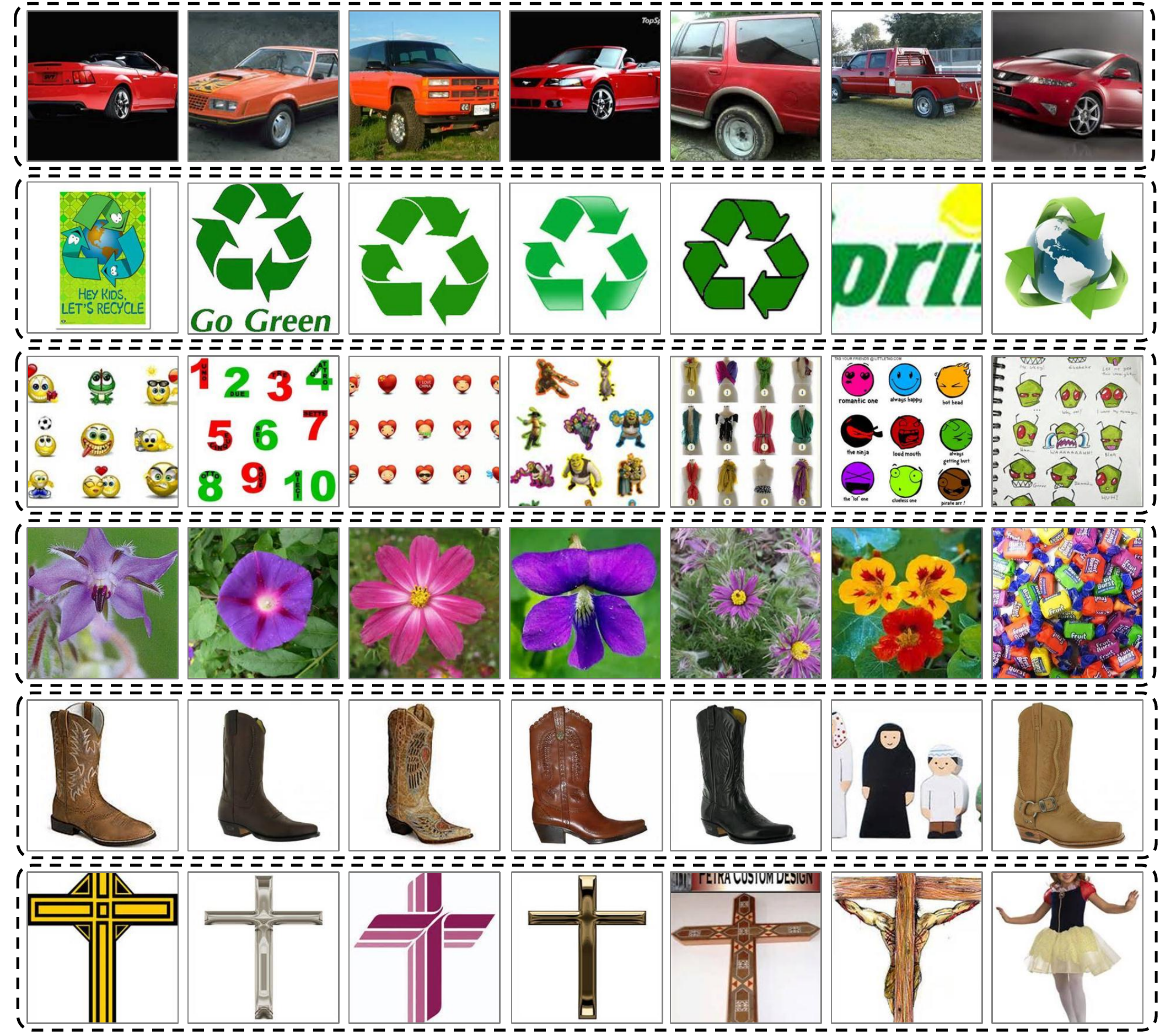}
\caption{Visualization of the learned common vector space by six randomly selected dimensions. The top 7 images with highest response at each dimension are showed in each row. It can be seen that clear semantic meanings or visual patterns are captured by these dimensions.}\label{mapping}
\end{figure*}

\begin{figure*}[t]
\centering
\includegraphics[width=0.75\textwidth,page=1]{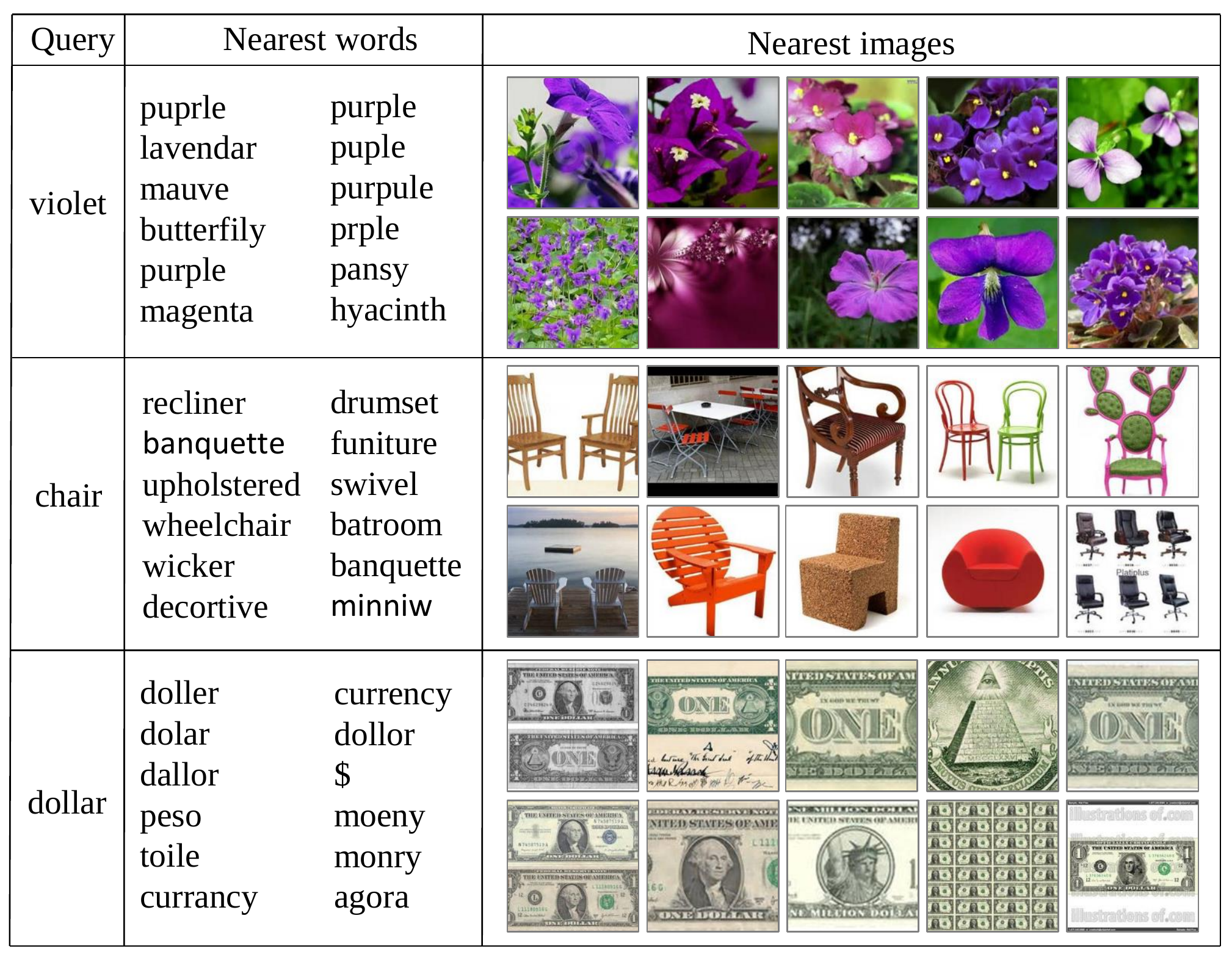}
\caption{Nearest neighbors of some exemplar words in the learned mapping space. Column 1 contains three query words, Column 2 and column 3 are their nearest words and images in feature space measured by the similarity defined by inner product.}\label{Fig:WbasedKNN}
\end{figure*}

Firstly, we qualitatively demonstrate the effectiveness of CSM by visualizing the learned mapping space.
Figure~\ref{mapping} visualizes six randomly selected dimensions of the learned feature space, and the images with high responses of each dimension are showed. The pattern captured by each dimension is both visually and semantically meaningful. Figure~\ref{Fig:WbasedKNN} demonstrates the effectiveness of inner product in the learned common space by showing nearest neighbor words and images of some exemplar words measured by inner products. Though the nearest words contains some spelling mistakes, it is easy to guess the real meaning.

\subsection{Overall Performance}
In order to validate the overall performance of CSM, we compare CSM with the two state-of-the-art single models on the dataset, i.e., Concept Classification model~\cite{carneiro2005formulating} and Passive-Aggressive model~\cite{grangier2008discriminative}.
Concept Classification model builds a binary classifier for each concept using a standard SVM.
Passive-Aggressive model utilizes a parametric function to map image into text space, and optimizes a learning criterion related to ranking performance.
Both Concept Classification model and Passive-Aggressive model adopt HOG features as the representation of images.
In addition, we compare our result with the ideal ranker and the random ranker.
The ideal ranking is the optimal ranking list generated by the relevance annotated by annotators, and the random result is a random order of the candidate images.
As mentioned in section~\ref{evaCr}, $\textrm{DCG}_{25}$ is utilized to capture quantitative results of the performance of the ranking list.
The overall performance is shown in Table~\ref{tab_overall}.

Because of the nature of test dataset, the average $\textrm{DCG}_{25}$ of ideal ranking is less than 1, since excellent candidates image for some test queries are less than 25. CSM achieves much better results than the state-of-the-arts models using sophisticate hand-craft images features, which quantitatively demonstrate the effectiveness of CSM in measure image-similarity for image retrieval.

\begin{table}
\centering
\normalsize
\caption{Overall performance of CSM.}
\begin{tabular}{|c|c|}
\hline
Model & Average $\textrm{DCG}_{25}$ \\
\hline
Random Ranking & 0.468 \\
Concept Classification model~\cite{carneiro2005formulating} & 0.494 \\
Passive-Aggressive model~\cite{grangier2008discriminative}  & 0.496 \\
CSM & 0.505 \\
Ideal Ranking & 0.684 \\
\hline
\end{tabular}
\label{tab_overall}
\end{table}

\subsection{Detailed Results Analysis}
In the whole 1,000 test queries, 71 queries achieve ideal retrieval performance and other 235 queries' DCG$_{25}$ are inferior to ideal ranking within 0.05.

\begin{figure*}[t]
\centering
\includegraphics[width=0.9\textwidth,page=1]{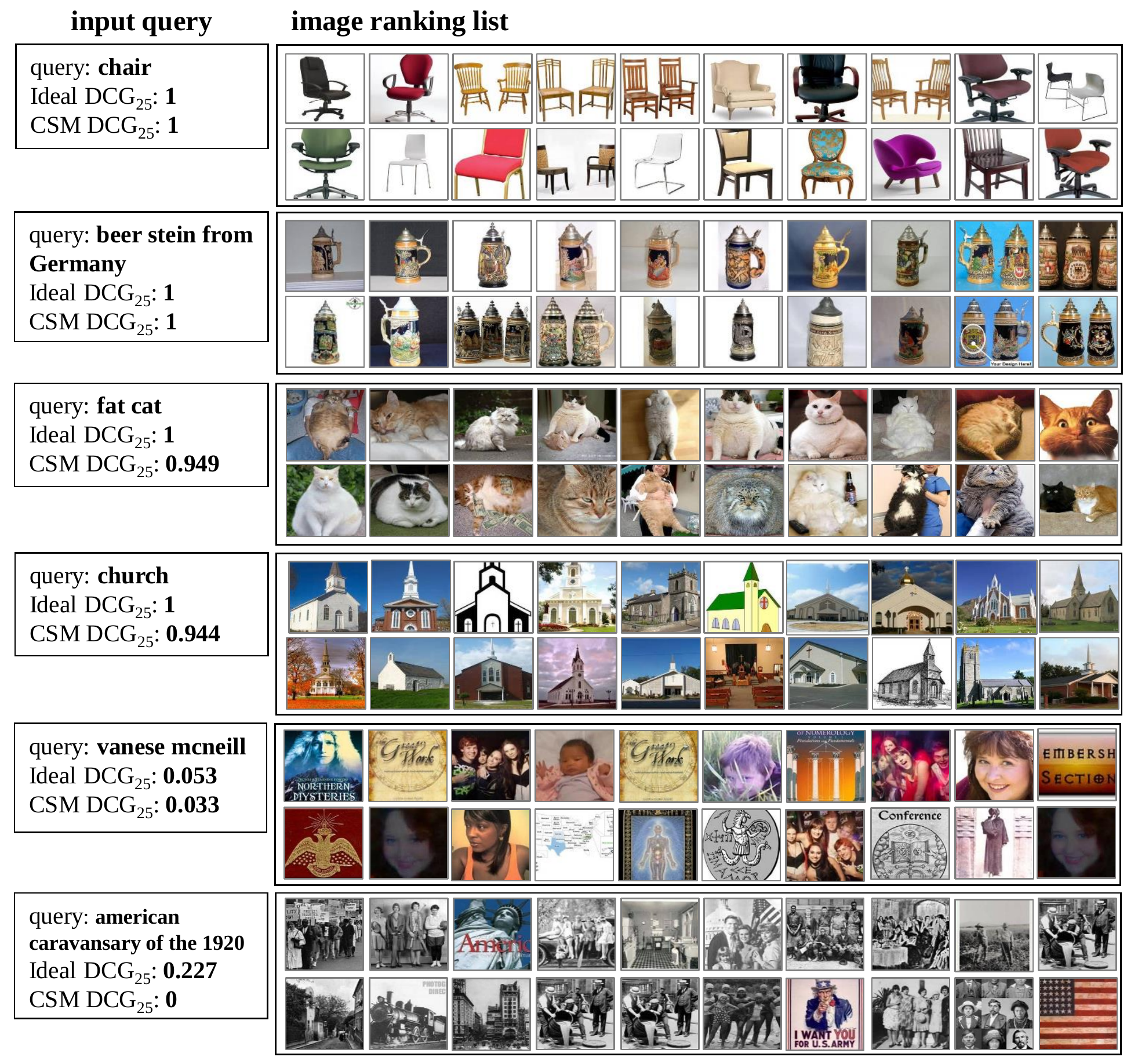}
\caption{Search results of several queries.}\label{Fig:Example}
\end{figure*}

Figure~\ref{Fig:Example} shows six retrieval ranking results by CSM including four queries achieved DCG$_{25}$ above 0.9 and two failure cases with DCG$_{25}$ close to 0.
The query \emph{chair}, \emph{fat cat} and \emph{church} can be matched exactly within the training query set.
Though the query \emph{beer stein from Germany} can not be matched exactly, the training query \emph{beer stein} is helpful to map effective textual weight vector through WEN.
The first failure case of \emph{vanese mcneill} is caused by the fact of ideal DCG$_{25}$ nearly being zero, since there are rare relevant images for this query.
For the last failure case of \emph{american caravansary of the 1920}, the key word \emph{caravansary} is missing in training word set, while other words are not helpful to map effective textual weight vector.

\begin{figure}[ht]
\centering
\includegraphics[width=0.48\textwidth,page=1]{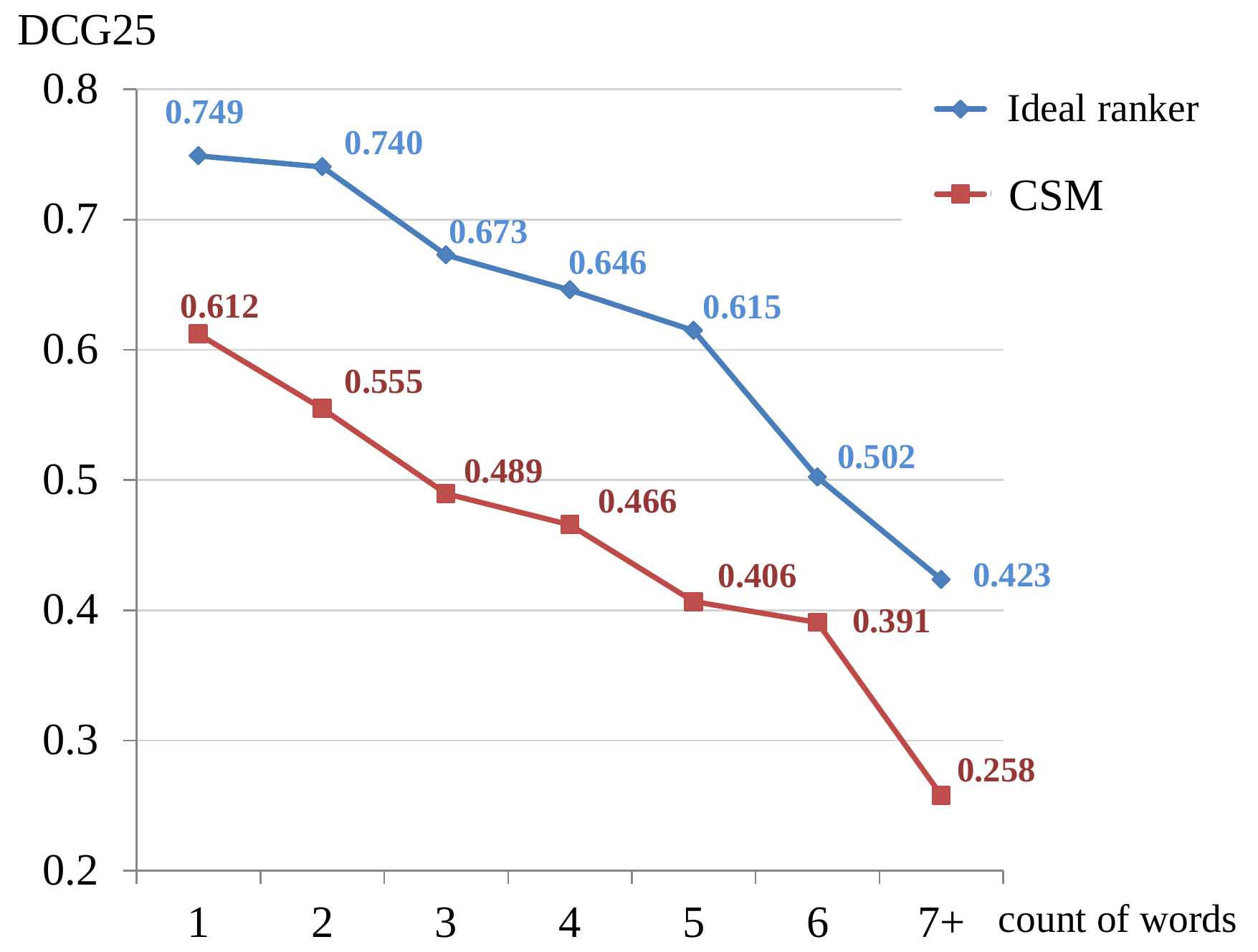}
\caption{Retrieval performance with different query lengths.}\label{Length}
\end{figure}

In addition, we further discuss the effects of retrieval performance vs the query length.
On one hand, more words embellish the search intention and limit the number of available candidate images, which is demonstrated by Figure~\ref{Length}, where longer queries are with lower $\text{DCG}_{25} score with ideal ranker$.

\begin{figure}[ht]
\centering
\includegraphics[width=0.48\textwidth,page=1]{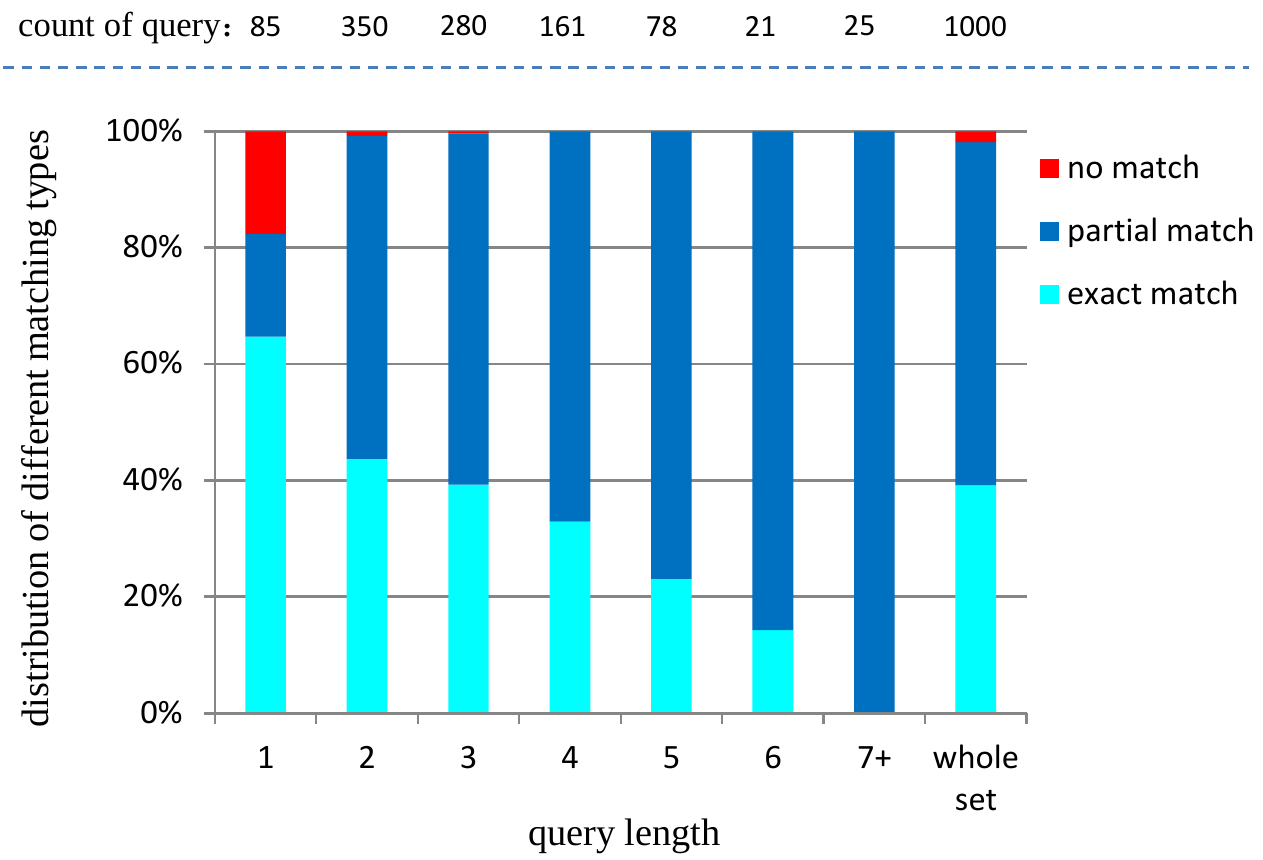}
\caption{The distribution of matching types with different query lengths and the statistics of query with different query lengths.}\label{ratioImg}
\end{figure}

On the other hand, different query lengths are with different query matching types.
Statistically, 392 test queries have exactly matched queries in the training query set, while 19 test queries have no matched queries in the training set.
The left 589 test queries can be partly matched, which means these queries contains one or more words in training query set.
Different matching types lead to different ranking performance, where no match leads to random ranker as previous mentioned and exact match can produce better ranking results.
The partial match is likely to introduce the semantic ambiguous, since the queries with partial match usually are matched with several training queries.
As shown in Figure~\ref{ratioImg}, the longer query set includes higher proportion of test queries with partial match.

\section{Conclusions}
\label{Conclusion}
In this paper, we proposed a novel approach for image retrieval, which reformulates image retrieval problem as mapping images and textual queries to one common space with an unified deep neural network. With sufficient training image provided by user clicks, the trained DNN significantly improved the image retrieval performance compared with state-of-the-arts methods based on predefined image features. In addition, CSM model not only measures the similarity between query and image, but also measures the similarity of textual queries and the similarity of images.
As the query embedding part still affects by the out of vocabulary problem, in the future we will combine the word embedding from natural language process tasks to enhance query embedding.

\bibliographystyle{IEEEtranS}
\bibliography{CSM}

\end{document}